\def\year{2019}\relax
\documentclass[letterpaper]{article} 
\usepackage{aaai19}  
\usepackage{times}  
\usepackage{helvet} 
\usepackage{courier}  
\usepackage[hyphens]{url}  
\usepackage{graphicx} 
\def\UrlFont{\rm}  
\usepackage{graphicx}  
\usepackage{todonotes}
\usepackage{gensymb}

\usepackage[utf8]{inputenc}

\title{Towards A Robot Explanation System: A Survey and Our Approach to\\ State Summarization, Storage and Querying, and Human Interface}

\author{
Zhao Han\\Department of Computer Science\\University of Massachusetts Lowell\\1 University Ave, Lowell, MA
\And
Jordan Allspaw\\Department of Computer Science\\University of Massachusetts Lowell\\1 University Ave, Lowell, MA
\AND
Adam Norton\\New England Robotics Validation and Experimentation Center\\University of Massachusetts Lowell\\1 University Ave, Lowell, MA
\And
Holly A. Yanco\\NERVE Center and CS Dept\\University of Massachusetts Lowell\\1 University Ave, Lowell, MA
}

\begin{document}

\maketitle

\begin{abstract}
As robot systems become more ubiquitous, developing understandable robot systems becomes increasingly important in order to build trust. In this paper, we present an approach to developing a holistic robot explanation system, which consists of three interconnected components: state summarization, storage and querying, and human interface. To find trends towards and gaps in the development of such an integrated system, a literature review was performed and categorized around those three components, with a focus on robotics applications. After the review of each component, we discuss our proposed approach for robot explanation. Finally, we summarize the system as a whole and review its functionality.
\end{abstract}

\section{Introduction}

With the advancement and wide adoption of deep learning techniques, explainability of software systems and interpretability of machine learning models has attracted both human-computer interaction (HCI) researchers (e.g., \cite{abdul2018hci}) and the artificial intelligence (AI) community (e.g., \cite{miller2019}). Work in human-robot interaction (HRI) has shown that improving understanding of a robot makes it more trustworthy \cite{desai2013trust} and more efficient \cite{admoni2016nonverbal}. However, how robots can explain themselves at a holistic level (i.e., generate explanations and communicate them, with a supporting data storage system with efficient querying) remains an open research question.

As opposed to virtual AI agents or computer software, robots have physical embodiment, which influences metrics such as empathy \cite{seo2015sorry} and cooperation \cite{bainbridge2008} with humans. Given this embodiment, some research in human-agent interaction is not applicable to human-robot interaction. For example, in a literature review about explainable agents and robots \cite{anjomshoae2019explainable}, approximately half ($47\%$) of the explanation systems examined used text-based communication methods, which is less relevant for robots that are not usually equipped with display screens. Instead, HRI researchers have been exploring  non-verbal physical behavior such as arm movement \cite{dragan2013legibility,kwon2018expressing} and eye gaze \cite{moon2014meet}. Non-verbal behaviors can help people to anticipate a robot’s actions \cite{lasota2017survey}, but understanding why that behavior occurred can improve one's prediction of behaviors, especially if the behavior is opaque \cite{malle2006book}. Thus, robot explanations of their own behavior are needed.

\begin{figure}[t]
\centering
\includegraphics[width=1\columnwidth]{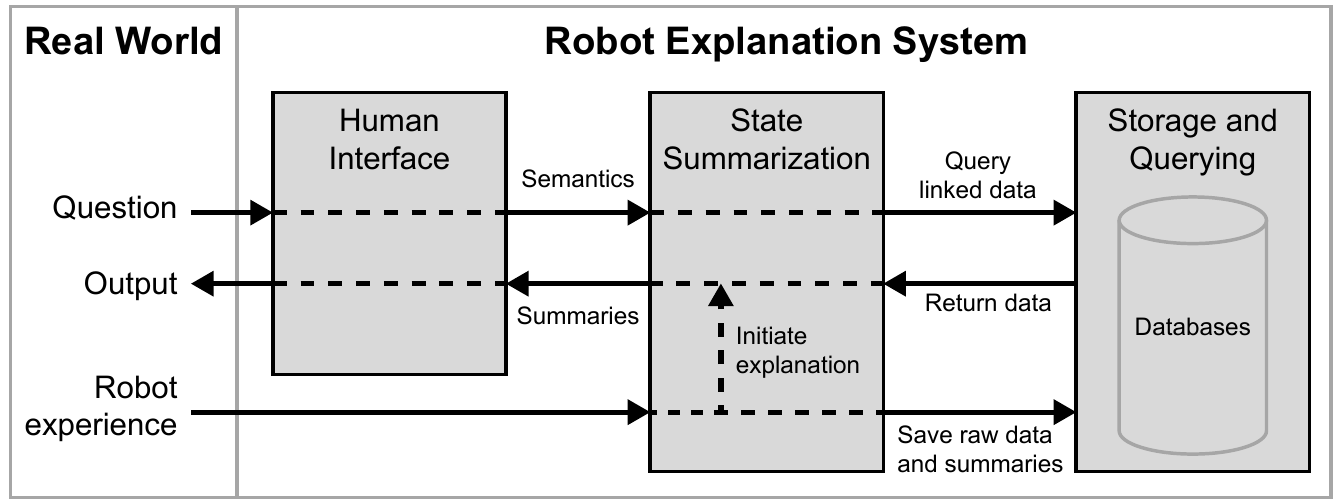} 
\caption{High-level representation of the robot explanation system's three components. Fig. \ref{fig:diagram} shows a detailed version.}
\label{fig:overview}
\end{figure}

In this paper, aholistic robot explanation system is decomposed into three components (see Figure \ref{fig:overview}), the research literature is surveyed to explore trends and gaps, and a proposed designed philosophy and approach is detailed as informed by the results of the literature review. We aim to provide important considerations and directions towards a robot explanation system that can accelerate robot acceptance. The term ``summarization'' addresses the process of shortening the description of the robot's activities while ``explanation'' strives to give insight into why the robot performed the summarized behaviors.

\subsection{System Components}

A robot explanation system requires state summarization, data storage and querying, and a human interface.

State summarization is at the core of the system, manually or automatically generating varying levels of summaries from different robot states while performing tasks or from the stored states in a post-hoc fashion. The varying summary levels allow people to receive explanations ranging from more abstract to more detailed \cite{brooks2010state} (e.g., processed data compared to raw sensor data, respectively). Explanations that utilize raw sensor data will likely only be favored by the minority of expert users while explanations involving processed data will be useful for both expert users and the majority of non-expert users.

A persistent storage system is needed to retain robot data and generated explanations. This system will pass the generated explanations, or summaries, to the human interface to be communicated. While storing them, different levels of explanations stemming from the same instance of source data need to be linked to maintain fluid interactions with a person. The person may request follow up explanations with more or less detail than the initial explanation. The storage system must also have a query component as part of the database interface to support online state summarization, which is needed when the stored summaries are not sufficient to answer users' questions. Querying must also be efficient, given the potentially large amount of robot data being stored.

The human interface component communicates the explanations from the robot to the human and allows the person to ask the system questions. The human interface can use several different modalities, such as natural language dialogues, a traditional graphic user interface (GUI) on a display screen or virtual reality (VR), and augmented reality (AR) that directly projects onto the robot's environment. The communication method could also involve moving the robot system, such as moving the robot's head or arm.

\subsection{Scope and Contributions of the Work}

This paper surveys the literature about the three components of an explanation system to provide critique, summarize trends, and discover gaps towards the design of a robot explanation system. By leveraging the trends observed in the research literature, we propose a robot explanation system architecture that will aim to fill the discovered gaps. 

The literature review was focused on research involving robots, rather than more general AI agents or the interpretation of machine learning models. This constraint was relaxed when work with robots was underrepresented or shared commonalities in one component, such as state summarization where abstract states both apply to AI agents and robots.

Thus, the literature review is not meant to be exhaustive, but comprehensive under these constraints to cover the three components in the context of physical co-located robot systems. We refer interested readers to \cite{abdul2018hci} for explainability of computer software, \cite{miller2019,adadi2018peeking} for the explainability of virtual AI agents, and \cite{zhang2018visual,guidotti2018survey} for interpretability of machine learning/deep learning models.

\section{State Summarization}

Before a robot can begin to explain its actions, it must first translate its decisions in a manner that could be understood by a human. Significant research has been performed within the fields of HCI and HRI towards this goal. In this section, we discuss systems in the literature that are deployable to physical robots. There has been significant research in the field of explainable AI, sometimes referred to as XAI \cite{wang2017residual}; however, this is beyond our scope. We are specifically interested in summarization methods that can work for a variety of different systems.

\subsection{Manual Methods}

There are two components of state summarization: the state of which the robot is aware and the state that the robot communicates to the user. A common approach is for developer to manually create categories by which the robot can explain its actions. For example, programmer specified function annotations for each designated robot action are used in \cite{hayes2017improving}. By creating a set of robot actions, correlated with code functions, the system is able to snapshot the state of the robot before and after a function is called. Since the state of the robot could be exceedingly large in a real world, deployed system, the state space is shrunk by isolating which variables are predetermined to be most relevant. These annotated variables are recorded every time a pre- and post-action snapshot is made. The robot then uses inspection to compare the pre- and post-variables of one action, compared to other similar successful actions, to make judgments.

A different approach, suggested in \cite{kaptein2017personalised}, is to adapt hierarchical task analysis \cite{schraagen2000cognitive} to a goal hierarchy tree (GHT). This involves creating a tree where the top node would be a high level task, which can be broken into a number of sub-goals, each linked by a belief (i.e., condition). Each sub-goal can then be broken into either sub-goals or actions. Choosing one sub-goal or action over another is based on a belief. The GHT can then be used to generate explanations. When comparing goal based vs. belief based explanations, \citeauthor{kaptein2017personalised} found that adults significantly preferred goal based explanations.

\subsection{Summarization Algorithms}

While manually creating categories or explanations can be effective, it is time consuming and not easily generalizable. Many techniques attempt to automate the process.

Programmer supplied explanations might be able to accurately describe the state of a robot, however, they can prove to be inadequate for a user. \citeauthor{ehsan2019automated} state that it is best to use a rationale justification \shortcite{ehsan2019automated} to explain to non-expert users, differentiating between a rationale and an explanation. An explanation can be made by exposing the inner workings of a system, but this type of explanation may not be understandable from non-experts. They suggest the alternative, a rationale, is meant to be an accessible and intuitive way of describing what the robot is doing. They also discuss how explanations can be tailored to optimize for different factors, including relatability, intelligibility, contextual accuracy, awareness and strategic detail; these factors can affect the user's confidence, understandability of the explanations, and how human-like explanation was. The approach does not attempt to provide an explanation that reveals the underlying algorithm, but rather attempts to justify an action based on how a non-developer bystander would think. The authors explore two different explanation strategies: ``focused view rationale" provides concise and localized rationale, which is more intelligible, and easier to understand, whereas ``complete view rationale" provides detailed and holistic rationale, which has better strategic detail and increased awareness.

\citeauthor{haidarian2010metacognitive} proposed a metacognitive loop (MCL) architecture with a generalized metacognition module that monitors and controls the performance of the system  \shortcite{haidarian2010metacognitive}. Every decision performed by the system has a set of expectations and a set of corrections or corrective responses. Their framework does not attempt to monitor and respond to specific expectation failures which would require intricate knowledge of how the world works. However, the abandonment of intricate knowledge makes it difficult to provide specialized, highly detailed explanations to an expert operator.

Most of this prior work examined explanations within rule-based and logic-based AI systems, not addressing the quantitative nature of much of the AI used in HRI. More recent work on automatic explanations instead used Partially Observable Markov Decision Problems (POMDPs) which have seen success in several situations within robotics \cite{wang2016impact}. Unfortunately, the quantitative nature of these models and the complexity of their solution algorithms also makes POMDP-reasoning opaque to people. \citeauthor{wang2016impact} propose an approach to automatically generate natural-language explanations for POMDP-based reasoning, with predefined string representations of the potential actions, accompanied by the level of uncertainty, and the relative likelihood of outcomes. The system could also reveal information about its sensing abilities along with how accurate its sensor is likely to be. However, modeling using POMPDs can be time consuming.

\citeauthor{miller2019} discusses how explanations delivered to the user should be generated based on data from social and behavioral research, which could increase user understandability \shortcite{miller2019}. Whether the explanation is generated from expert developers or from a large dataset of novice operators, both cases still require manually tying the robot algorithm to an explanation, a process that can be difficult and faulty.

In the literature review by \citeauthor{anjomshoae2019explainable}, they conclude that context-awareness and personalization remain under-researched despite having been determined to be key factors in explainable agency \shortcite{anjomshoae2019explainable}. They also suggest that multi-model explanation presentation is possibly useful, which would mean your underlying state representation would need to be robust enough to handle several different approaches. Finally, they propose that a robot should keep track of a user's knowledge, with the explanation generation model updated to reflect the evolution of user expertise.
    
\subsection{Proposed Approach}
Much of the prior research focuses on scenarios where the system is designed for either novice or expert users but not both. \citeauthor{de2017people} argue that a robot should take a person's knowledge and role into account when formulating a response \shortcite{de2017people}. While context-awareness and personalization have been outlined as key factors for effective state summarization, there is little research where the robot adjusts its explanation depending on the user \cite{anjomshoae2019explainable}, even in simple cases such as expert vs. novice user.

There are many identified differences between expert users and novice users. For example, traceability and verification are very important for software and hardware engineers \cite{cleland2012software} while explainability or intelligibility are particularly important for laymen \cite{de2017people}. Cases where a user starts as a novice then gains experience over time are under-researched. In our system, the state representation and explanation generation need to be able to adjust to the user and possibly change depending on context. In addition, even if the user is taken into account, such situations fail to account for cases where a user could potentially want to have both levels of explanations available simultaneously or to switch between. For example, The user could quickly receive a high level explanation for why the robot performed an action, then if that proves insufficient, inquire for more details. Ideally, the robot would be able to provide explanations with varying granularity and context, tailored to the experience level of the user (e.g., bystander, operator, programmer, etc.).

Given the conclusion that participants preferred annotations of the actions by a reinforcement learning game agent from developers \cite{ehsan2019automated}, both manual and automated methods to generate explanations should be considered. Manually generated explanations fill the gap that new users do not have a thorough understanding of the logic in the underlying algorithm. However, when developers manually put explanations in code (e.g., using \cite{hayes2017improving}), one should always consider the new users audience and provide easy-to-understand explanations that are not tightly coupled with implementation details.

In order to cover a wide variety of possible situations, our proposed approach is for an expertly created inner state representation based on categories and goals with methods to automatically create desirable explanations delivered to the user, where those explanations can adjust based on the user's experience (bystander, novice, expert) as well as other factors. The system should isolate and convey necessary context for a decision or state, or be prepared to provide it if additional information is requested. Specifically, if confidence in these generated responses in low, the state summarization algorithms can fall back to the expertly created explanations.

For example, the sample task of picking up an object could be decomposed into subtasks: locate the object, navigate to it, then grasp the located object. Each of those subtasks can then be broken down to subtasks of their own. The last step of grasping can be decomposed to reaching, grabbing, and retreating arm back to home location. Eventually the subtasks end up in robot primitives, the simplest actions the system can describe. Each of these actions can have a failure reason, along with context, which would include sensor data and prior relevant state information. If the motion planner fails to find a valid solution, the error propagates up and the action of ``grasping an object'' failed because the subtask ``grabbing the object'' failed when ``reach for the object'' failed as a result of no valid inverse kinematics solution being found. The system needs to correctly determine the most relevant and useful failure level to report; for example, in this case, an expert operator would be told, ``No valid inverse kinematics solution was found'' while a bystander would be told ``I could not reach the object.''

\section{Storage and Querying}

Terminal output or logs are common methods for debugging during active development and for error analysis after a robot has been deployed, but both methods have some drawbacks. Terminal output is essentially volatile memory, lost after the terminal window is closed, disallowing retrospection. However, despite being persistent on disks, software logs are unstructured and unlinked between related data, which makes it hard to effectively and efficiently query. Thus, researchers have been exploring database techniques to better store and query robotic data. Because storage and querying are under-discussed in the robotic community, this section is more detailed than the other two components.

\subsection{Storing Unprocessed Data} 

Many researchers have been leveraging the schemaless MongoDB database to store unprocessed data from sensors or communication messages from lower-level middleware such as motion planners \cite{niemueller2012genericdb,beetz2015openease}. Being schemaless allows for recording different hierarchical data messages without declaring the hierarchy in the database (i.e., tables in relational databases such as MySQL). One such hierarchical example is the popular Pose message type present in the Robot Operating System (ROS) framework \cite{ros}. A Pose message contains a position Point message and an orientation Quaternion message; a Point message contains float values $x$, $y$, and $z$; an orientation message is represented by $x$, $y$, $z$, and $w$. It is imaginable to go through the cumbersome process of creating tables of Pose, Point, and Quaternion. Even more tables have to be created for each hierarchical data message. This advantage is also described as minimal configuration and allows evolving data structures to support innovation and development \cite{niemueller2012genericdb}.

Niemueller, Lakemeyer and Srinivasa open-sourced the $mongodb\_log$ library and are among the first to introduce MongoDB to robotics for logging purposes, which has applications to fault analysis and performance evaluation \cite{niemueller2012genericdb}. In addition to being schemaless, the features that support scalability, such as capped collections, indexing and replication, are discussed. Capped collections handles limited storage capability by replacing old records with new ones. Indexing on a field or a combination of fields speeds up querying. Replication allows storing data across computers using the distributed pragma. Note that the indexing and replication features are also supported by relational databases.

While low-level data is needed, recording all raw data will soon hit the storage capacity limit: when old data is replaced by new records, the important information in the old data will be lost. This is particularly true when the data comes at a high rate; e.g., a HERB robot generates 0.1 GB per minute typically and 0.5 GB at peak times \cite{niemueller2012genericdb}. A more effective way is to be selective, only storing the data of interest \cite{oliveira2014perceptual}. However, storing raw sensor data only facilitates debugging for developers; it does not solve the high-level explanation storage that will help non-expert users to understand the robot.

In addition, while it might be appropriate to expose the database to developers, a more effective way may be an interface that hides the database complexity, easing the cognitive burden on developers. This could be programming language agnostic, for example, by having a HTTP REST API.

Other researchers have also used MongoDB to store low-level data \cite{beetz2010cram,niemueller2013towards,winkler2014cramm,balint2017storing} except for \citeauthor{oliveira2014perceptual} \shortcite{oliveira2014perceptual}  who used LevelDB, a key-value database for perceived object data. \citeauthor{ravichandran2018workbench} benchmarked major types of databases and found on average MongoDB has the best performance to continuous robotic data \shortcite{ravichandran2018workbench}. However, time-series and key-value databases are not included in the benchmark.

\subsection{Storing Processed Data}

Instead of looking for related data using the universal time range, \citeauthor{balint2017storing} proposed Common Analysis Structure to store linked data for manipulation tasks \shortcite{balint2017storing}. The structure includes timestamp, scene, image, and camera information. A scene has a viewpoint coordinate frame, annotations, and object hypotheses. Annotations are supporting planes or a semantic location, and object hypotheses are regions of raw data and their respective annotations. The authors considered storage space constraints, thus filtering and storing only regions of interest in unblurred images or point clouds. In their follow-up work \cite{durner2017experience}, the Common Analysis Structure is used to optimize perception parameters by users providing ground truth labels.

Similarly, \citeauthor{oliveira2014perceptual} proposed a perception database using LevelDB to enable object category learning from users \shortcite{oliveira2014perceptual}. Instead of regions of raw point cloud data, user mediated key views of the same object are stored linking to one object category.

\citeauthor{wang2012cloudrobot} utilized a relational database as cloud robotics storage so multiple low-end robots can retrieve 3D laser scan data from a high-end robot, which has a laser sensor and its data being processed onboard with more storage and better computation power \shortcite{wang2012cloudrobot}. Specifically, low-end robots can send a query with their poses on a map to retrieve 3D map data and image data. PostgreSQL is used but the data structure detail is not discussed, as the paper focuses on resource allocation and scheduling. However, a local data buffer on robots is proposed to store frequently accessed data to reduce the database access latency bottleneck.

Dietrich et al. used Cassandra to store and query 2D and 3D map data with spatial context such as building, floor, and room \shortcite{dietrich2014distributed}. There are several benefits of using Cassandra, such as the ability to have a local server that can query both local data and remote data, avoiding single point failure. Developers can also define TTLs (Time to Live) to remove data automatically, avoiding a maintenance burden.

In addition, \citeauthor{fourie2017slamindb} leveraged a graph database, Neo4j, to link vision sensor data stored in MongoDB to pose-keyed data \shortcite{fourie2017slamindb}. Graph databases allow complex queries with spatial context for multiple mapping mobile robots, which enables multi-robot mapping. This line of research focuses on storing processed data but did not discuss a way to link raw data back to the processed data. This is important because not storing linked raw data may lead to loss of information during retrospection. There is a trend that other types of database systems, e.g., relational database (PostgreSQL) and key-value database (LevelDB and Cassandra) are used to store those processed data, because only a few ever-evolving data structures need to be stored.

\subsection{Querying}

There is no unified method for querying; most are application specific, such as efficient debugging \cite{niemueller2012genericdb,balint2017storing} and task representation \cite{beetz2015openease,tenorth2015openease}. Interfaces are also tightly coupled to programming languages: JavaScript from MongoDB \cite{niemueller2012genericdb}, Prolog \cite{beetz2015openease,tenorth2015openease} and SQL \cite{dietrich2015selectscript}.

In $mongodb\_log$, Niemueller, Lakemeyer and Srinivasa proposed a knowledge hierarchy for manipulation tasks to enable efficient querying for debugging \shortcite{niemueller2012genericdb}. The knowledge hierarchy consists of all raw data and the poses of the robot and manipulated objects, all of which are timestamped. When a manipulation task fails, a top-down search is performed in the knowledge hierarchy in a specific time range. Poses are at the root of the hierarchy and raw data, such as coordinate frames and point cloud data, are replayed in a visualization tool for further investigation (i.e., Rviz in ROS). The query language is JavaScript using the MapReduce paradigm, which supports aggregation of data natively.

\citeauthor{beetz2015openease} proposed Open-EASE, a web interface for robotic knowledge representation and processing for developers \cite{beetz2015openease,tenorth2015openease}. Robotics and AI researchers are able to encapsulate manipulation tasks semantically as temporal events with sets of predefined semantic predicates. Manipulation episodes are logged by storing low-level data, which are the environment model, object detection results and poses, and planned tasks in an XML-based Web Ontology Language (OWL) \cite{owl}. High-velocity raw data such as sensor data and robot poses are logged in a schemaless MongoDB database. Querying uses Prolog with a predefined concept vocabulary, similar to the semantic predicates.

While Open-EASE allows semantic querying, it does not come easily. One disadvantage is the introduction of a different programming paradigm, logic programming in Prolog, which robot developers have to learn for querying regardless of the paradigm being used for robot programming. It is also unclear how to extend the pre-defined semantic predicates for other generic tasks in different environments.

\citeauthor{balint2017storing} use a similar high level description language to replace the JavaScript query feature in MongoDB \shortcite{balint2017storing} to avoid the in-depth knowledge requirement of the internal data structure. The description language also contains predefined predicates and can be queried through Prolog. This work has the same drawbacks as Open-EASE.

Interestingly, \citeauthor{dietrich2015selectscript} proposed SelectScript, a SQL-inspired query-only language for robotic world models without having relevant tables in the database \shortcite{dietrich2015selectscript}. Without using a different programming language to specify how to retrieve data, SelectScript provides a declarative and language-agnostic way to specify what data are needed rather than how. SelectScript also features custom function support to queries and custom return type native to robotic applications such as an occupancy grid map.

While SelectScript is modeled on the well-known standard SQL, but it is not language-agnostic as stated. Custom functions are only supported in Python, leaving ROS C++ programmers behind. Except for requiring significant effort to support C++, it is not trivial to extend return type to new data types such as the popular Octomap used in 3D mapping \cite{hornung13auro} for obstacle avoidance in SelectScript.

\citeauthor{fourie2017slamindb} proposed to use a graph database to query spatial data from multiple mobile robots \shortcite{fourie2017slamindb}. However, it is not plausible for our use given that only one relationship is used: odometry poses linked to image and RGBD data. This work also suffers the same drawbacks of SelectScript in that custom queries have to be programmed in Java.

Similar to our argument in the previous storage sections, robotic database designers should embrace the programming languages with which robotic developers are already familiar. Database technology should be hidden by interfaces written in programming languages that also support access to the underlying database for advanced and customized use.

\subsection{Proposed Approach}

MongoDB's use has been proven by robotic developers and should thus be chosen to store low-level sensor data, which is potentially large and high velocity. This is mainly to replace the rosbag utility\footnote{http://wiki.ros.org/rosbag} which relies on a filesystem and is not easy to query. To store summaries at different levels and to link them, a relational database should be chosen because it is specialized to store relational data. Additional columns will be used as reference to the sensor data in MongoDB. When data are deleted in MongoDB or the relational database, the linked data should be deleted as well; this can be achieved by a background job system or enforced in the programming interface to be discussed below.

Instead of directly exposing the database to robot developers, the storage and querying interface should be written in the same programming language the developer is using for the robot system. The programming interface should be written in an object-oriented manner to allow easy extension (e.g., from a single robot to a cluster). A minimum subset of functional programming should also be used to support custom functions, similar to SelectScript \cite{dietrich2015selectscript}. For common use cases, the determination of which data storage method to use should be handled by the interface so users do not need to be concerned with the underlying database being used. However, a raw interface that enables developers to directly communicate with the underlying database should not be completely left out, to allow for use cases that are not in consideration by interface designers. In terms of programming languages, C++ should be used, given its popularity among developers and most packages of the ROS framework. Python support should also be provided through binding the C++ implementation\footnote{http://wiki.ros.org/ROS/Tutorials/Using a C++ class in Python}.

Human interface developers should be able to query using indexes such as state summarization level, time range, and others specific to the domain. Custom queries should also be allowed by having interface functions exposing the database, as interface designers are not able to consider an exhaustive list of all use cases. Human interface developers should also be able to easily query linked follow-up data after the first query. This is essential for the human interface to provide interaction (e.g., interactive conversation or projection).

\section{Human Interface}

A human interface is used to communicate the explanations generated by the robot. Communication of the explanations can occur in different channels, such as a traditional graphical user interface (GUI) on a monitor, head-mounted displays, and robot movements. While some human interface methods have been studied for decades in the HCI community, our literature review is selective. We focus largely on novel approaches (e.g., AR techniques) and the most prominent related work, justified by the citation number relative to the publication year. There is a large body of research for some techniques with existing comprehensive literature reviews. We direct readers to the following papers: eye gaze in social robotics \cite{admoni2017eyegaze}, using animation techniques with robots \cite{schulz2019animation} (which provides 12 design guidelines), speech and natural language processing for robotics \cite{mavridis2015review}, and tactile communication via artificial skins in social robots \cite{silvera2015artificial}.

\subsection{Display Screen}

While computer interfaces largely use a display screen for the primary communication channel, screens on robots are largely used to display facial expressions \cite{kalegina2018characterizing} due to their physicality, but are considered less convenient than speech \cite{de2018towards}. For co-location scenarios, it is rare to find a display screen as part of a robot that is not attached to its head, so very little research has been performed for simple displays or visualizations of sensor values or other relevant information.

\citeauthor{brooks2017human} investigated displaying a general set of state icons on the body of robots to indicate internal states \shortcite{brooks2017human}. Five icons -- OK, Help, Off, Safe, and Dangerous -- were shown to participants for evaluation. The results show that while bystanders are able to understand those icons, their level of understanding is vague. For example, the "Off" icon could be interpreted as stating that the robot is powered off or that it is just not currently actively operating.

SoftBank Robotics' Pepper robot is one of the only robot systems that features a touch screen not attached to its head. \citeauthor{feingold9differences} found that people enjoyed interacting with a touch screen on a robot more than using a computer screen with a keyboard \shortcite{feingold9differences}. Specifically, participants preferred to use the touch screen to indicate the completion of a task. \citeauthor{de2018towards} used Pepper's screen to present buttons to use for inputting instructions, such as object directions \shortcite{de2018towards}. \citeauthor{bruno2018culturally} used Pepper's touch screen like a tablet where multiple-choice questions are shown and users can answer by tapping on the choices \shortcite{bruno2018culturally}.

While a display screen has been demonstrated to be effective at showing accurate information (e.g., replaying past events \cite{jeong2017study}, which can be used during explanations), there is sometimes a mental conversion issue where humans have to map what is displayed on the screen to the physical environment. A display screen may also suffer from being less readable from a longer distance, which is important as such proximity to a robot may not be safe during certain failure cases \cite{honig2018understanding}.

\subsection{Augmented Reality (AR)}

Utilizing AR for explainability allows visual cues to be projected directly into the environment with which the robot interacts, allowing for more specificity and reference points to be drawn. This technique can make explanations more accurate, less ambiguous, and remove the burden of mental mapping between different reference frames (e.g., 2D display screen compared to the real world 3D environment).

\citeauthor{andersen2016projecting} proposed to use a projector to communicate a robot's intent and task information onto the workspace to facilitate human-robot collaboration in a manufacturing environment \shortcite{andersen2016projecting}. The robot locates a physical car door using an edge-based detection method, then projects visualizations of parts onto it to indicate its perception and intended manipulation actions. The authors also conducted an experiment by asking participants to collaboratively rotate and move cubes with the robot arm, comparing the AR projector method to the use of a display screen with text. Results show there were fewer performance errors and questions asked by the participants when using the projector method.

For mobility, researchers also leveraged projection techniques onto the ground to show robot intention. \citeauthor{chadalavada2015s} projected a green line to indicate the planned path and two white lines to the left and right of the robot to visualize the collision avoidance range of the robot \shortcite{chadalavada2015s}. Gradient light bands have also been used to show a robot's path \cite{watanabe2015communicating}. Similarly, \citeauthor{coovert2014spatial} projected arrows to show the robot's path \shortcite{coovert2014spatial}, while \citeauthor{daily2003world} used a head-mounted display to visualize the robot's path onto the user's view of the environment \shortcite{daily2003world}. Circles have also been used to show landmarks on a robot swarm using a projector located above the performance space \cite{ghiringhelli2014interactive}. However, the AR techniques utilize in these papers are passive and not interactive. \citeauthor{chakraborti2018projection} proposed using Microsoft HoloLens to enable a user to interact with AR projections \shortcite{chakraborti2018projection}, where users can use pinch gestures to move a robot's arm or base, start or stop robot movement, and pick a block for stacking.

AR may be more salient than a display screen for our use case, but it does have some drawbacks. For example, it cannot be used for a robot to take initiative for explanations: This is because it can be easily ignored if a human is not paying attention to the projected area.

\subsection{Robot Activities}

\begin{figure*}[t]
\centering
\includegraphics[width=0.95\textwidth]{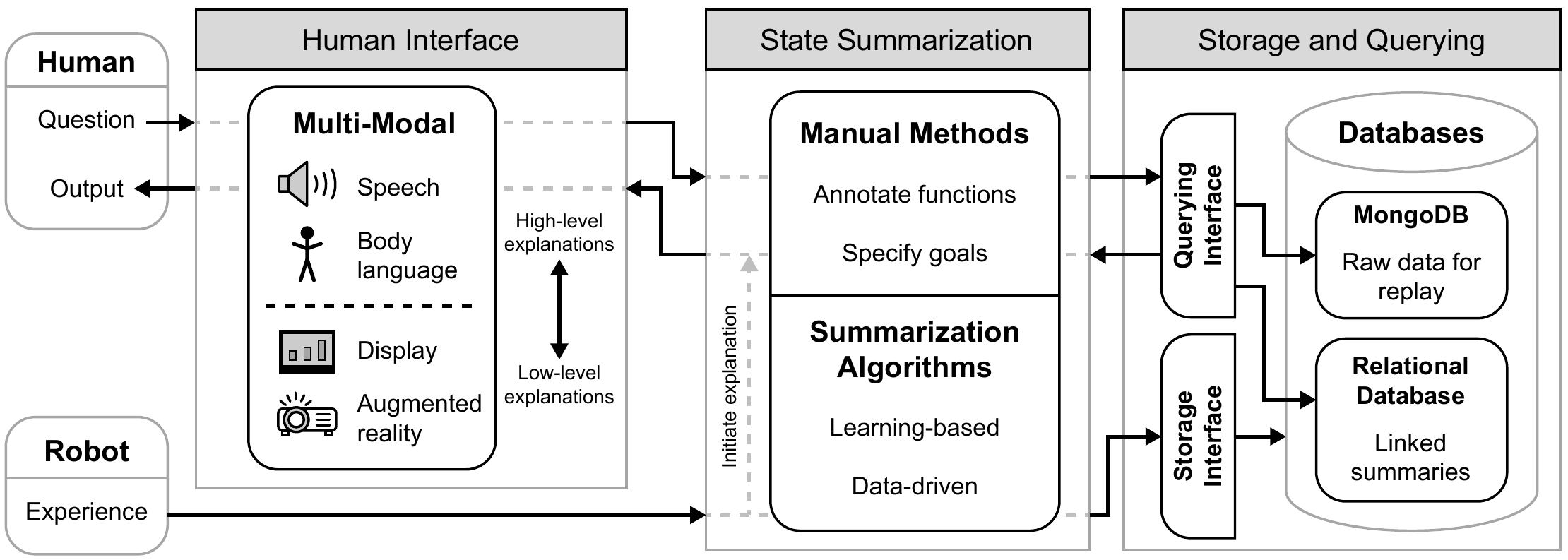} 
\caption{The workflow of the robot explanation system. State summaries are saved in the databases which are hidden by programming interfaces. A multi-modal human interface queries the database through another programming interface and answers follow-up conversation and interactions. The state summarization component can also initiate explanations.}
\label{fig:diagram}
\end{figure*}

Due to the physicality of robot systems, body language of robots has been studied extensively in the HRI community to communicate intent. For example, \citeauthor{dragan2015effects} proposed using legible robot arm movements to allow people to quickly infer the robot's next grasp target \shortcite{dragan2015effects}. Repeated arm movement has also been proposed to communicate a robot's incapability to pick up an object \cite{kwon2018expressing}. Eye gaze behavior or head movement has also been studied (e.g., \cite{moon2014meet,admoni2017eyegaze}). However, this communication method is limited in the amount of information that it can convey, if used as the only channel of communication.

In addition to robot movements, researchers have also explored auxiliary methods of communication such as light. Notably, the Rethink Robotics Baxter system utilizes a ring of lights on its head to indicate the distance of humans moving nearby to support safe HRI. Similarly, \citeauthor{szafir2015communicating} used light to indicate the flying direction of a drone when co-located with humans in close proximity \shortcite{szafir2015communicating}; the results show improvement in response time and accuracy.

\subsection{Proposed Approach}

Given that speech is a natural interaction method for people, it should be considered for initial explanations. It can be used to garner a person's attention and to initiate high-level summary explanations. However, it is limited in that only one audio stream is available at a time (i.e., the human cannot listen to multiple streams simultaneously without interference). Body language can be used to supplement speech explanations, but likely should not be used as the only communication channel. For example, robots can use arm movements to refer to relevant geography of the robot (e.g., components, actions) and the task space (e.g., objects, areas of the environment) simultaneously with another communication channel such as speech.

When a human requests more detailed explanations, other communication channels should be used to avoid misinterpretation. One such interface is the AR projection method in the literature, given the limitations of display screens (i.e., availability, size, reading distance). While using a projection method, one should keep in mind that projection on ground is not always visible \cite{chakraborti2018projection}.

Communication of explanations may occur in the three temporal levels --- a priori, in situ, or post hoc --- which will impact the effectiveness of a chosen communication channel. More detailed, in-depth communications via visual or audio means may be better suited for explanations of planned actions (a priori) or analysis of resulting actions (post hoc). Simpler techniques for alerting people (e.g., flashing lights, vibrating tactors) may be better suited for conveying state information in situ, at least to garner attention before more information is conveyed.

Thus, communication should be multi-modal as some methods are better suited for different levels of explanations, temporal levels, and data types, but also need to ensure the human understands all possible means of communication.

\section{Integration}

Next we describe the integration of the three components of our approach for and their interconnections in the robot explanation system. The proposed component designs are intended to serve as guidelines for development with reasonable justifications, rather than compulsory decisions. In addition, the design of each component should be self-contained and decoupled from the operation of other components, allowing each component to be used independently.

\subsection{System Review and Workflow}

A diagram of the system is shown in Figure \ref{fig:diagram}. There are two main methods for state summarization: manual methods and summarization algorithms. Manually, developers can annotate functions or specify goals as leaves in a tree structure in the robotic applications to provide explanations. When using summarization algorithms, explanations can be learned using end-to-end or semi-supervised deep learning from robot states and annotator-provided explanations. Summarization algorithms should also be able generalize summaries online when the stored summaries are not sufficient to answer some users' questions. While generating explanations, the state summarization component can initiate explanations if necessary (e.g., in cases of incapability or failures).

The generated state summaries and their linked raw data are then saved to databases through a programming interface, currently using C++ or Python. Two databases are used: MongoDB for storing raw sensor data and a relational database to store explanations (i.e., linked summaries). The two data storage methods are mostly hidden by the interface, allowing developers to use the programming language and avoid knowing database details. However, the interface will also provide ways to directly access each database if needed.

With the stored summaries in the database, after the robot or human initiates communication, the human interface is responsible for all follow-up conversations or interactions by passing the semantics to the state summarization component. Communication should occur in multiple modalities including speech, body language, screen, and AR (e.g., projection techniques). Due to the differences of each communication method in terms of fidelity, attention-getting, etc., the system will utilize multi-modal communication.

\subsection{Evaluation}

To evaluate the effectiveness of system, usability testing should be performed with users of varying experience to assess the acceptance and understanding of the system.

\begin{figure}[t]
\centering
\includegraphics[width=0.9\columnwidth]{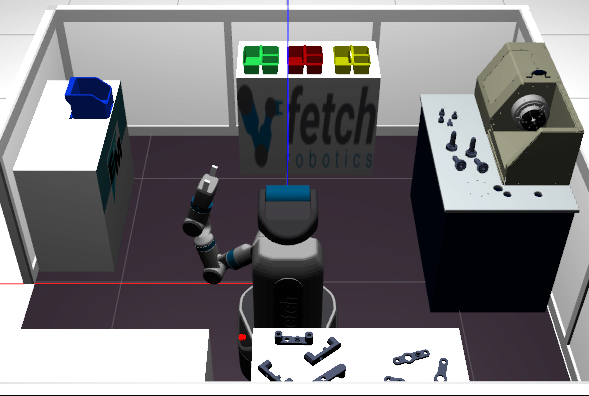} 
\caption{The FetchIt! task environment we recommend for evaluation. The robot's goal is to place the irregular parts into the correct sections of the concave caddy, and transport the caddy to the bottom-left table for inspection.}
\label{fig:fetchit}
\end{figure}

For an example scenario and tasks, we recommend using the FetchIt! mobile manipulation challenge \cite{fetchitws}; Figure \ref{fig:fetchit} shows a rendering of the task space. The tasks are to assemble a kit by navigating to collect parts on different station; while scoped for a manufacturing environment, the same types of tasks are relevant to home environments: e.g., navigating between areas in a narrow hallway kitchen and a dining table, and manipulating objects in these places. The challenges are also not singular to a work cell manufacturing environment. One challenge is detecting different objects, whose shapes are complex and irregular (e.g., large bolts and gearbox parts; kitchen utensils would be similarly difficult). Invisible from the rendered task space, the screws are in the blue container. Another challenge is to navigate through the narrow and constrained work cell; the available space and obstructions therein are similar to that of a kitchen.

The FetchIt! environment provides a reasonable test bed for a robot explanation system because there are several opportunities for unexpected or opaque events to occur. For example, a common occurrence is that the Fetch robot may not be able to grasp a caddy or a gearbox part if it is placed too close to a wall. Fetch's arm may not be long enough to reach given the constraints presented by the end-effector orientation (it must be pointed down in order to grasp the caddy) and standoff distances imposed by the dimensions of the tables. These scenarios are not apparent to novice users or bystanders who do not have intimate knowledge of Fetch's characteristics. In this scenario, Fetch should initiate an explanation to inform the user. Another common occurrence is confusion when differentiating between two gearbox parts that appear similar in height via point cloud due to sensor noise. This can make the object detection fail, causing the robot to grasp the incorrect object. In this scenario, a human might initiate a robot explanation, as the robot may not be aware that it performed incorrectly. Another human-initiated robot explanation might be when the robot stops at a different location in front of the caddy table than what was expected, and places a part into the incorrect caddy. This could occur due to navigation error range and the narrow horizontal field of view (54\degree) of its RGBD camera, which may cause part of the caddy to be occluded. 

The FetchIt! competition testbed is available for Gazebo on GitHub: \url{https://github.com/fetchrobotics/fetch_gazebo/tree/gazebo9/fetchit_challenge}. A working implementation, including navigation and manipulation, is available at \url{https://github.com/uml-robotics/fetchit2019}.

\section{Future Work}

To date, we have started implementing the robot explanation system. The system will be open-sourced to facilitate advancing research and to assist other practitioners in integrating their existing software with the system. We plan to evaluate the implementation with a formal HRI user study and analyze the results for further improvements.

\section{Conclusion}

This paper presents a survey of the three components of the robot explanation system. For state summarization, manually generated summaries may be the workaround solution due to lack of maturity of learning methods that required more research effort. For storage and querying, the programming interface should be developed for easy integration from state summarization and human interface developers. Multi-modal human interface communication methods should be used not only to garner attention from humans in initiating an explanation, but also as the enabling methods to convey both high level summarized explanations and low level detailed explanations.

\section*{Acknowledgements}
This work has been supported in part by the Office of Naval Research (N00014-18-1-2503).

\bibliographystyle{aaai}
\bibliography{bib}

\begin{thebibliography}{}

\bibitem[\protect\citeauthoryear{Abdul \bgroup et al\mbox.\egroup
  }{2018}]{abdul2018hci}
Abdul, A.; Vermeulen, J.; Wang, D.; Lim, B.~Y.; and Kankanhalli, M.
\newblock 2018.
\newblock Trends and trajectories for explainable, accountable and intelligible
  systems: An hci research agenda.
\newblock In {\em Proceedings of the 2018 CHI Conference on Human Factors in
  Computing Systems},  582:1--582:18.

\bibitem[\protect\citeauthoryear{Adadi and Berrada}{2018}]{adadi2018peeking}
Adadi, A., and Berrada, M.
\newblock 2018.
\newblock Peeking inside the black-box: A survey on explainable artificial
  intelligence (xai).
\newblock {\em IEEE Access} 6:52138--52160.

\bibitem[\protect\citeauthoryear{Admoni and
  Scassellati}{2017}]{admoni2017eyegaze}
Admoni, H., and Scassellati, B.
\newblock 2017.
\newblock Social eye gaze in human-robot interaction: a review.
\newblock {\em Journal of Human-Robot Interaction} 6(1):25--63.

\bibitem[\protect\citeauthoryear{Admoni \bgroup et al\mbox.\egroup
  }{2016}]{admoni2016nonverbal}
Admoni, H.; Weng, T.; Hayes, B.; and Scassellati, B.
\newblock 2016.
\newblock Robot nonverbal behavior improves task performance in difficult
  collaborations.
\newblock In {\em The Eleventh ACM/IEEE International Conference on Human Robot
  Interaction},  51--58.

\bibitem[\protect\citeauthoryear{Andersen \bgroup et al\mbox.\egroup
  }{2016}]{andersen2016projecting}
Andersen, R.~S.; Madsen, O.; Moeslund, T.~B.; and Amor, H.~B.
\newblock 2016.
\newblock Projecting robot intentions into human environments.
\newblock In {\em 2016 25th IEEE International Symposium on Robot and Human
  Interactive Communication (RO-MAN)},  294--301.

\bibitem[\protect\citeauthoryear{Anjomshoae \bgroup et al\mbox.\egroup
  }{2019}]{anjomshoae2019explainable}
Anjomshoae, S.; Najjar, A.; Calvaresi, D.; and Fr{\"a}mling, K.
\newblock 2019.
\newblock Explainable agents and robots: Results from a systematic literature
  review.
\newblock In {\em Proceedings of the 18th International Conference on
  Autonomous Agents and MultiAgent Systems},  1078--1088.
\newblock International Foundation for Autonomous Agents and Multiagent
  Systems.

\bibitem[\protect\citeauthoryear{Bainbridge \bgroup et al\mbox.\egroup
  }{2008}]{bainbridge2008}
Bainbridge, W.~A.; Hart, J.; Kim, E.~S.; and Scassellati, B.
\newblock 2008.
\newblock The effect of presence on human-robot interaction.
\newblock In {\em The 17th IEEE International Symposium on Robot and Human
  Interactive Communication (RO-MAN)},  701--706.

\bibitem[\protect\citeauthoryear{{Balint-Benczédi} \bgroup et al\mbox.\egroup
  }{2017}]{balint2017storing}
{Balint-Benczédi}, F.; {Márton}, Z.; {Durner}, M.; and {Beetz}, M.
\newblock 2017.
\newblock Storing and retrieving perceptual episodic memories for long-term
  manipulation tasks.
\newblock In {\em 2017 18th International Conference on Advanced Robotics
  (ICAR)},  25--31.

\bibitem[\protect\citeauthoryear{Beetz, M{\"o}senlechner, and
  Tenorth}{2010}]{beetz2010cram}
Beetz, M.; M{\"o}senlechner, L.; and Tenorth, M.
\newblock 2010.
\newblock {CRAM} — a cognitive robot abstract machine for everyday
  manipulation in human environments.
\newblock In {\em 2010 IEEE/RSJ International Conference on Intelligent Robots
  and Systems},  1012--1017.
\newblock IEEE.

\bibitem[\protect\citeauthoryear{Beetz, Tenorth, and
  Winkler}{2015}]{beetz2015openease}
Beetz, M.; Tenorth, M.; and Winkler, J.
\newblock 2015.
\newblock {Open-EASE} — a knowledge processing service for robots and
  robotics/ai researchers.
\newblock In {\em 2015 IEEE International Conference on Robotics and Automation
  (ICRA)},  1983--1990.

\bibitem[\protect\citeauthoryear{Brooks \bgroup et al\mbox.\egroup
  }{2010}]{brooks2010state}
Brooks, D.; Shultz, A.; Desai, M.; Kovac, P.; and Yanco, H.~A.
\newblock 2010.
\newblock Towards state summarization for autonomous robots.
\newblock In {\em 2010 AAAI Fall Symposium Series}.

\bibitem[\protect\citeauthoryear{Brooks}{2017}]{brooks2017human}
Brooks, D.~J.
\newblock 2017.
\newblock {\em A Human-Centric Approach to Autonomous Robot Failures}.
\newblock Ph.D. Dissertation, Ph. D. dissertation, Department of Computer
  Science, University.

\bibitem[\protect\citeauthoryear{Bruno \bgroup et al\mbox.\egroup
  }{2018}]{bruno2018culturally}
Bruno, B.; Menicatti, R.; Recchiuto, C.~T.; Lagrue, E.; Pandey, A.~K.; and
  Sgorbissa, A.
\newblock 2018.
\newblock Culturally-competent human-robot verbal interaction.
\newblock In {\em 2018 15th International Conference on Ubiquitous Robots
  (UR)},  388--395.

\bibitem[\protect\citeauthoryear{Chadalavada \bgroup et al\mbox.\egroup
  }{2015}]{chadalavada2015s}
Chadalavada, R.~T.; Andreasson, H.; Krug, R.; and Lilienthal, A.~J.
\newblock 2015.
\newblock That's on my mind! robot to human intention communication through
  on-board projection on shared floor space.
\newblock In {\em 2015 European Conference on Mobile Robots (ECMR)},  1--6.

\bibitem[\protect\citeauthoryear{Chakraborti \bgroup et al\mbox.\egroup
  }{2018}]{chakraborti2018projection}
Chakraborti, T.; Sreedharan, S.; Kulkarni, A.; and Kambhampati, S.
\newblock 2018.
\newblock Projection-aware task planning and execution for human-in-the-loop
  operation of robots in a mixed-reality workspace.
\newblock In {\em 2018 IEEE/RSJ International Conference on Intelligent Robots
  and Systems (IROS)},  4476--4482.
\newblock IEEE.

\bibitem[\protect\citeauthoryear{Cleland-Huang \bgroup et al\mbox.\egroup
  }{2012}]{cleland2012software}
Cleland-Huang, J.; Gotel, O.; Zisman, A.; et~al.
\newblock 2012.
\newblock {\em Software and systems traceability}, volume~2.
\newblock Springer.

\bibitem[\protect\citeauthoryear{Coovert \bgroup et al\mbox.\egroup
  }{2014}]{coovert2014spatial}
Coovert, M.~D.; Lee, T.; Shindev, I.; and Sun, Y.
\newblock 2014.
\newblock Spatial augmented reality as a method for a mobile robot to
  communicate intended movement.
\newblock {\em Computers in Human Behavior} 34:241--248.

\bibitem[\protect\citeauthoryear{Daily \bgroup et al\mbox.\egroup
  }{2003}]{daily2003world}
Daily, M.; Cho, Y.; Martin, K.; and Payton, D.
\newblock 2003.
\newblock World embedded interfaces for human-robot interaction.
\newblock In {\em 36th Annual Hawaii International Conference on System
  Sciences, 2003. Proceedings of the},  6--pp.
\newblock IEEE.

\bibitem[\protect\citeauthoryear{De~Graaf and Malle}{2017}]{de2017people}
De~Graaf, M.~M., and Malle, B.~F.
\newblock 2017.
\newblock How people explain action (and autonomous intelligent systems should
  too).
\newblock In {\em 2017 AAAI Fall Symposium Series}.

\bibitem[\protect\citeauthoryear{de Jong \bgroup et al\mbox.\egroup
  }{2018}]{de2018towards}
de~Jong, M.; Zhang, K.; Roth, A.~M.; Rhodes, T.; Schmucker, R.; Zhou, C.;
  Ferreira, S.; Cartucho, J.; and Veloso, M.
\newblock 2018.
\newblock Towards a robust interactive and learning social robot.
\newblock In {\em Proceedings of the 17th International Conference on
  Autonomous Agents and MultiAgent Systems},  883--891.

\bibitem[\protect\citeauthoryear{Desai \bgroup et al\mbox.\egroup
  }{2013}]{desai2013trust}
Desai, M.; Kaniarasu, P.; Medvedev, M.; Steinfeld, A.; and Yanco, H.
\newblock 2013.
\newblock Impact of robot failures and feedback on real-time trust.
\newblock In {\em Proceedings of the 8th ACM/IEEE International Conference on
  Human-robot Interaction},  251--258.

\bibitem[\protect\citeauthoryear{Dietrich \bgroup et al\mbox.\egroup
  }{2014}]{dietrich2014distributed}
Dietrich, A.; Zug, S.; Mohammad, S.; and Kaiser, J.
\newblock 2014.
\newblock Distributed management and representation of data and context in
  robotic applications.
\newblock In {\em 2014 IEEE/RSJ International Conference on Intelligent Robots
  and Systems},  1133--1140.
\newblock IEEE.

\bibitem[\protect\citeauthoryear{Dietrich, Zug, and
  Kaiser}{2015}]{dietrich2015selectscript}
Dietrich, A.; Zug, S.; and Kaiser, J.
\newblock 2015.
\newblock Selectscript: A query language for robotic world models and
  simulations.
\newblock In {\em 2015 IEEE International Conference on Robotics and Automation
  (ICRA)},  6254--6260.

\bibitem[\protect\citeauthoryear{Dragan \bgroup et al\mbox.\egroup
  }{2015}]{dragan2015effects}
Dragan, A.~D.; Bauman, S.; Forlizzi, J.; and Srinivasa, S.~S.
\newblock 2015.
\newblock Effects of robot motion on human-robot collaboration.
\newblock In {\em Proceedings of the Tenth Annual ACM/IEEE International
  Conference on Human-Robot Interaction},  51--58.
\newblock ACM.

\bibitem[\protect\citeauthoryear{Dragan, Lee, and
  Srinivasa}{2013}]{dragan2013legibility}
Dragan, A.~D.; Lee, K.~C.; and Srinivasa, S.~S.
\newblock 2013.
\newblock Legibility and predictability of robot motion.
\newblock In {\em Proceedings of the 8th ACM/IEEE international conference on
  Human-robot interaction},  301--308.

\bibitem[\protect\citeauthoryear{Durner \bgroup et al\mbox.\egroup
  }{2017}]{durner2017experience}
Durner, M.; Kriegel, S.; Riedel, S.; Brucker, M.; M{\'a}rton, Z.-C.;
  B{\'a}lint-Bencz{\'e}di, F.; and Triebel, R.
\newblock 2017.
\newblock Experience-based optimization of robotic perception.
\newblock In {\em 2017 18th International Conference on Advanced Robotics
  (ICAR)},  32--39.

\bibitem[\protect\citeauthoryear{Ehsan \bgroup et al\mbox.\egroup
  }{2019}]{ehsan2019automated}
Ehsan, U.; Tambwekar, P.; Chan, L.; Harrison, B.; and Riedl, M.~O.
\newblock 2019.
\newblock Automated rationale generation: a technique for explainable ai and
  its effects on human perceptions.
\newblock In {\em Proceedings of the 24th International Conference on
  Intelligent User Interfaces (IUI)},  263--274.

\bibitem[\protect\citeauthoryear{Feingold-Polak \bgroup et al\mbox.\egroup
  }{2018}]{feingold9differences}
Feingold-Polak, R.; Elishay, A.; Shahar, Y.; Stein, M.; Edan, Y.; and
  Levy-Tzedek, S.
\newblock 2018.
\newblock Differences between young and old users when interacting with a
  humanoid robot: a qualitative usability study.
\newblock {\em Paladyn, Journal of Behavioral Robotics} 9(1):183--192.

\bibitem[\protect\citeauthoryear{fet}{}]{fetchitws}
Fetchit!, a mobile manipulation challenge.
\newblock \url{https://opensource.fetchrobotics.com/competition}.
\newblock Accessed: 2019-09-11.

\bibitem[\protect\citeauthoryear{Fourie \bgroup et al\mbox.\egroup
  }{2017}]{fourie2017slamindb}
Fourie, D.; Claassens, S.; Pillai, S.; Mata, R.; and Leonard, J.
\newblock 2017.
\newblock Slamindb: Centralized graph databases for mobile robotics.
\newblock In {\em 2017 IEEE International Conference on Robotics and Automation
  (ICRA)},  6331--6337.

\bibitem[\protect\citeauthoryear{Ghiringhelli \bgroup et al\mbox.\egroup
  }{2014}]{ghiringhelli2014interactive}
Ghiringhelli, F.; Guzzi, J.; Di~Caro, G.~A.; Caglioti, V.; Gambardella, L.~M.;
  and Giusti, A.
\newblock 2014.
\newblock Interactive augmented reality for understanding and analyzing
  multi-robot systems.
\newblock In {\em 2014 IEEE/RSJ International Conference on Intelligent Robots
  and Systems},  1195--1201.

\bibitem[\protect\citeauthoryear{Guidotti \bgroup et al\mbox.\egroup
  }{2018}]{guidotti2018survey}
Guidotti, R.; Monreale, A.; Ruggieri, S.; Turini, F.; Giannotti, F.; and
  Pedreschi, D.
\newblock 2018.
\newblock A survey of methods for explaining black box models.
\newblock {\em ACM computing surveys (CSUR)} 51(5):93.

\bibitem[\protect\citeauthoryear{Haidarian \bgroup et al\mbox.\egroup
  }{2010}]{haidarian2010metacognitive}
Haidarian, H.; Dinalankara, W.; Fults, S.; Wilson, S.; Perlis, D.; Schmill, M.;
  Oates, T.; Josyula, D.; and Anderson, M.
\newblock 2010.
\newblock The metacognitive loop: An architecture for building robust
  intelligent systems.
\newblock In {\em PAAAI Fall Symposium on Commonsense Knowledge
  (AAAI/CSK’10)}.

\bibitem[\protect\citeauthoryear{Hayes and Shah}{2017}]{hayes2017improving}
Hayes, B., and Shah, J.~A.
\newblock 2017.
\newblock Improving robot controller transparency through autonomous policy
  explanation.
\newblock In {\em Proceedings of the 2017 ACM/IEEE international conference on
  human-robot interaction},  303--312.

\bibitem[\protect\citeauthoryear{Honig and
  Oron-Gilad}{2018}]{honig2018understanding}
Honig, S., and Oron-Gilad, T.
\newblock 2018.
\newblock Understanding and resolving failures in human-robot interaction:
  Literature review and model development.
\newblock {\em Frontiers in psychology} 9:861.

\bibitem[\protect\citeauthoryear{Hornung \bgroup et al\mbox.\egroup
  }{2013}]{hornung13auro}
Hornung, A.; Wurm, K.~M.; Bennewitz, M.; Stachniss, C.; and Burgard, W.
\newblock 2013.
\newblock {OctoMap}: An efficient probabilistic {3D} mapping framework based on
  octrees.
\newblock {\em Autonomous Robots}.
\newblock Software available at \url{http://octomap.github.com}.

\bibitem[\protect\citeauthoryear{Jeong \bgroup et al\mbox.\egroup
  }{2017}]{jeong2017study}
Jeong, S.-Y.; Choi, I.-J.; Kim, Y.-J.; Shin, Y.-M.; Han, J.-H.; Jung, G.-H.;
  and Kim, K.-G.
\newblock 2017.
\newblock A study on ros vulnerabilities and countermeasure.
\newblock In {\em Proceedings of the Companion of the 2017 ACM/IEEE
  International Conference on Human-Robot Interaction},  147--148.

\bibitem[\protect\citeauthoryear{Kalegina \bgroup et al\mbox.\egroup
  }{2018}]{kalegina2018characterizing}
Kalegina, A.; Schroeder, G.; Allchin, A.; Berlin, K.; and Cakmak, M.
\newblock 2018.
\newblock Characterizing the design space of rendered robot faces.
\newblock In {\em Proceedings of the 2018 ACM/IEEE International Conference on
  Human-Robot Interaction},  96--104.

\bibitem[\protect\citeauthoryear{Kaptein \bgroup et al\mbox.\egroup
  }{2017}]{kaptein2017personalised}
Kaptein, F.; Broekens, J.; Hindriks, K.; and Neerincx, M.
\newblock 2017.
\newblock Personalised self-explanation by robots: The role of goals versus
  beliefs in robot-action explanation for children and adults.
\newblock In {\em 2017 26th IEEE International Symposium on Robot and Human
  Interactive Communication (RO-MAN)},  676--682.
\newblock IEEE.

\bibitem[\protect\citeauthoryear{Kwon, Huang, and
  Dragan}{2018}]{kwon2018expressing}
Kwon, M.; Huang, S.~H.; and Dragan, A.~D.
\newblock 2018.
\newblock Expressing robot incapability.
\newblock In {\em Proceedings of the 2018 ACM/IEEE International Conference on
  Human-Robot Interaction},  87--95.

\bibitem[\protect\citeauthoryear{Lasota \bgroup et al\mbox.\egroup
  }{2017}]{lasota2017survey}
Lasota, P.~A.; Fong, T.; Shah, J.~A.; et~al.
\newblock 2017.
\newblock A survey of methods for safe human-robot interaction.
\newblock {\em Foundations and Trends{\textregistered} in Robotics}
  5(4):261--349.

\bibitem[\protect\citeauthoryear{Malle}{2006}]{malle2006book}
Malle, B.~F.
\newblock 2006.
\newblock {\em How the mind explains behavior: Folk explanations, meaning, and
  social interaction}.
\newblock MIT Press.

\bibitem[\protect\citeauthoryear{Mavridis}{2015}]{mavridis2015review}
Mavridis, N.
\newblock 2015.
\newblock A review of verbal and non-verbal human--robot interactive
  communication.
\newblock {\em Robotics and Autonomous Systems} 63:22--35.

\bibitem[\protect\citeauthoryear{Miller}{2019}]{miller2019}
Miller, T.
\newblock 2019.
\newblock Explanation in artificial intelligence: Insights from the social
  sciences.
\newblock {\em Artificial Intelligence} 267:1 -- 38.

\bibitem[\protect\citeauthoryear{Moon \bgroup et al\mbox.\egroup
  }{2014}]{moon2014meet}
Moon, A.; Troniak, D.~M.; Gleeson, B.; Pan, M.~K.; Zheng, M.; Blumer, B.~A.;
  MacLean, K.; and Croft, E.~A.
\newblock 2014.
\newblock Meet me where i'm gazing: how shared attention gaze affects
  human-robot handover timing.
\newblock In {\em Proceedings of the 2014 ACM/IEEE international conference on
  Human-robot interaction},  334--341.

\bibitem[\protect\citeauthoryear{Niemueller \bgroup et al\mbox.\egroup
  }{2013}]{niemueller2013towards}
Niemueller, T.; Abdo, N.; Hertle, A.; Lakemeyer, G.; Burgard, W.; and Nebel, B.
\newblock 2013.
\newblock Towards deliberative active perception using persistent memory.
\newblock In {\em Proc. IROS 2013 Workshop on AI-based Robotics}.

\bibitem[\protect\citeauthoryear{Niemueller, Lakemeyer, and
  Srinivasa}{2012}]{niemueller2012genericdb}
Niemueller, T.; Lakemeyer, G.; and Srinivasa, S.~S.
\newblock 2012.
\newblock A generic robot database and its application in fault analysis and
  performance evaluation.
\newblock In {\em 2012 IEEE/RSJ International Conference on Intelligent Robots
  and Systems},  364--369.

\bibitem[\protect\citeauthoryear{Oliveira \bgroup et al\mbox.\egroup
  }{2014}]{oliveira2014perceptual}
Oliveira, M.; Lim, G.~H.; Lopes, L.~S.; Kasaei, S.~H.; Tom{\'e}, A.~M.; and
  Chauhan, A.
\newblock 2014.
\newblock A perceptual memory system for grounding semantic representations in
  intelligent service robots.
\newblock In {\em 2014 IEEE/RSJ International Conference on Intelligent Robots
  and Systems},  2216--2223.

\bibitem[\protect\citeauthoryear{Quigley \bgroup et al\mbox.\egroup
  }{2009}]{ros}
Quigley, M.; Conley, K.; Gerkey, B.; Faust, J.; Foote, T.; Leibs, J.; Wheeler,
  R.; and Ng, A.~Y.
\newblock 2009.
\newblock {ROS: an open-source Robot Operating System}.
\newblock In {\em ICRA Workshop on Open Source Software}, ~5.

\bibitem[\protect\citeauthoryear{Ravichandran \bgroup et al\mbox.\egroup
  }{2018}]{ravichandran2018workbench}
Ravichandran, R.; Prassler, E.; Huebel, N.; and Blumenthal, S.
\newblock 2018.
\newblock A workbench for quantitative comparison of databases in multi-robot
  applications.
\newblock In {\em 2018 IEEE/RSJ International Conference on Intelligent Robots
  and Systems (IROS)},  3744--3750.

\bibitem[\protect\citeauthoryear{Schraagen, Chipman, and
  Shalin}{2000}]{schraagen2000cognitive}
Schraagen, J.~M.; Chipman, S.~F.; and Shalin, V.~L.
\newblock 2000.
\newblock {\em Cognitive task analysis}.
\newblock Psychology Press.

\bibitem[\protect\citeauthoryear{Schulz, Torresen, and
  Herstad}{2019}]{schulz2019animation}
Schulz, T.; Torresen, J.; and Herstad, J.
\newblock 2019.
\newblock Animation techniques in human-robot interaction user studies: A
  systematic literature review.
\newblock {\em ACM Transactions on Human-Robot Interaction (THRI)} 8(2):12.

\bibitem[\protect\citeauthoryear{Seo \bgroup et al\mbox.\egroup
  }{2015}]{seo2015sorry}
Seo, S.~H.; Geiskkovitch, D.; Nakane, M.; King, C.; and Young, J.~E.
\newblock 2015.
\newblock Poor thing! would you feel sorry for a simulated robot?: A comparison
  of empathy toward a physical and a simulated robot.
\newblock In {\em Proceedings of the Tenth Annual ACM/IEEE International
  Conference on Human-Robot Interaction},  125--132.

\bibitem[\protect\citeauthoryear{Silvera-Tawil, Rye, and
  Velonaki}{2015}]{silvera2015artificial}
Silvera-Tawil, D.; Rye, D.; and Velonaki, M.
\newblock 2015.
\newblock Artificial skin and tactile sensing for socially interactive robots:
  A review.
\newblock {\em Robotics and Autonomous Systems} 63:230--243.

\bibitem[\protect\citeauthoryear{Szafir, Mutlu, and
  Fong}{2015}]{szafir2015communicating}
Szafir, D.; Mutlu, B.; and Fong, T.
\newblock 2015.
\newblock Communicating directionality in flying robots.
\newblock In {\em 2015 10th ACM/IEEE International Conference on Human-Robot
  Interaction (HRI)},  19--26.

\bibitem[\protect\citeauthoryear{Tenorth \bgroup et al\mbox.\egroup
  }{2015}]{tenorth2015openease}
Tenorth, M.; Winkler, J.; Be{\ss}ler, D.; and Beetz, M.
\newblock 2015.
\newblock {Open-EASE}: A cloud-based knowledge service for autonomous learning.
\newblock {\em KI-K{\"u}nstliche Intelligenz} 29(4):407--411.

\bibitem[\protect\citeauthoryear{W3C}{2009}]{owl}
W3C.
\newblock 2009.
\newblock {OWL} 2 web ontology language document overview.
\newblock {W3C} recommendation, W3C.
\newblock http://www.w3.org/TR/2009/REC-owl2-overview-20091027/.

\bibitem[\protect\citeauthoryear{Wang \bgroup et al\mbox.\egroup
  }{2012}]{wang2012cloudrobot}
Wang, L.; Liu, M.; Meng, M. Q.-H.; and Siegwart, R.
\newblock 2012.
\newblock Towards real-time multi-sensor information retrieval in cloud robotic
  system.
\newblock In {\em 2012 IEEE International Conference on Multisensor Fusion and
  Integration for Intelligent Systems (MFI)},  21--26.

\bibitem[\protect\citeauthoryear{Wang \bgroup et al\mbox.\egroup
  }{2017}]{wang2017residual}
Wang, F.; Jiang, M.; Qian, C.; Yang, S.; Li, C.; Zhang, H.; Wang, X.; and Tang,
  X.
\newblock 2017.
\newblock Residual attention network for image classification.
\newblock In {\em Proceedings of the IEEE Conference on Computer Vision and
  Pattern Recognition},  3156--3164.

\bibitem[\protect\citeauthoryear{Wang, Pynadath, and
  Hill}{2016}]{wang2016impact}
Wang, N.; Pynadath, D.~V.; and Hill, S.~G.
\newblock 2016.
\newblock The impact of pomdp-generated explanations on trust and performance
  in human-robot teams.
\newblock In {\em Proceedings of the 2016 international conference on
  autonomous agents \& multiagent systems},  997--1005.
\newblock International Foundation for Autonomous Agents and Multiagent
  Systems.

\bibitem[\protect\citeauthoryear{Watanabe \bgroup et al\mbox.\egroup
  }{2015}]{watanabe2015communicating}
Watanabe, A.; Ikeda, T.; Morales, Y.; Shinozawa, K.; Miyashita, T.; and Hagita,
  N.
\newblock 2015.
\newblock Communicating robotic navigational intentions.
\newblock In {\em 2015 IEEE/RSJ International Conference on Intelligent Robots
  and Systems (IROS)},  5763--5769.

\bibitem[\protect\citeauthoryear{Winkler \bgroup et al\mbox.\egroup
  }{2014}]{winkler2014cramm}
Winkler, J.; Tenorth, M.; Bozcuoglu, A.~K.; and Beetz, M.
\newblock 2014.
\newblock Cramm--memories for robots performing everyday manipulation
  activities.
\newblock {\em Advances in Cognitive Systems} 3:47--66.

\bibitem[\protect\citeauthoryear{Zhang and Zhu}{2018}]{zhang2018visual}
Zhang, Q.-s., and Zhu, S.-C.
\newblock 2018.
\newblock Visual interpretability for deep learning: a survey.
\newblock {\em Frontiers of Information Technology \& Electronic Engineering}
  19(1):27--39.

\end{thebibliography}

\end{document}